\documentclass{sig-alternate-05-2015}

\usepackage{graphicx}
\usepackage{balance}  
\usepackage{verbatim}
\usepackage{tikz}
\usepackage[draft]{todonotes}   
\usepackage{framed}
\usepackage[]{algorithm2e}
\usepackage{multirow}
\usepackage{array}

\newcommand{\struct}[1]{\texttt{#1}}
\newcommand{\phrase}[1]{\textit{`#1'}}
\newcommand{\utterance}[1]{\textit{``#1''}}

\newcommand{\old}[1]{}

\newcommand{\squishlist}{
	\begin{list}{$\bullet$}
		{ \setlength{\itemsep}{0pt}
			\setlength{\parsep}{3pt}
			\setlength{\topsep}{3pt}
			\setlength{\partopsep}{0pt}
			\setlength{\leftmargin}{1.5em}
			\setlength{\labelwidth}{1em}
			\setlength{\labelsep}{0.5em} } }

	\newcommand{\squishlisttwo}{
		\begin{list}{$\bullet$}
			{ \setlength{\itemsep}{0pt}
				\setlength{\parsep}{0pt}
				\setlength{	opsep}{0pt}
				\setlength{\partopsep}{0pt}
				\setlength{\leftmargin}{2em}
				\setlength{\labelwidth}{1.5em}
				\setlength{\labelsep}{0.5em} } }
		
		\newcommand{\squishend}{
		\end{list}  }

\begin{document}

\setcopyright{rightsretained}

\doi{}

\isbn{}



%

\title{Knowledge Questions from Knowledge Graphs}
%
%
%
%
%

\numberofauthors{3} 
%
\author{
%
%
\alignauthor
Dominic Seyler$^{\ddagger}$\\
\affaddr{\vspace{2mm}$^{\ddagger}$University of Illinois at Urbana-Champaign}\\
\affaddr{Urbana, Illinois}\\
\email{\vspace{2mm}dseyler2@illinois.edu}
\alignauthor
Mohamed Yahya$^{\dagger}$\\
\affaddr{\vspace{2mm}$^{\dagger}$Max Planck Institute for Informatics}\\
\affaddr{Saarland Informatics Campus}\\
\email{\vspace{1mm}myahya@mpi-inf.mpg.de}
\alignauthor Klaus Berberich$^{*,\dagger}$\\
\affaddr{\vspace{2mm}$^{*}$htw saar}\\
\affaddr{Saarbr\"ucken, Germany}\\
\affaddr{\texttt{~}}\\
\email{\vspace{2mm}klaus.berberich@htwsaar.de}
}


\maketitle

\begin{abstract}
  We address the novel problem of automatically generating quiz-style
  knowledge questions from a knowledge graph such as
  DBpedia. Questions of this kind have ample applications, for
  instance, to educate users about or to evaluate their knowledge in a
  specific domain. To solve the problem, we propose an end-to-end
  approach. The approach first selects a named entity from the
  knowledge graph as an answer. It then generates a structured
  triple-pattern query, which yields the answer as its sole result. If
  a multiple-choice question is desired, the approach selects
  alternative answer options. Finally, our approach uses a
  template-based method to verbalize the structured query and yield a
  natural language question. A key challenge is estimating how
  difficult the generated question is to human users. To do this, we
  make use of historical data from the Jeopardy! quiz show and a
  semantically annotated Web-scale document collection, engineer
  suitable features, and train a logistic regression classifier to
  predict question difficulty. Experiments demonstrate the viability
  of our overall approach.
\end{abstract}

\section{Introduction} 
\label{sec:introduction} 

 \begin{figure}[t!]
 	\centering
 	\includegraphics[width=1\linewidth]{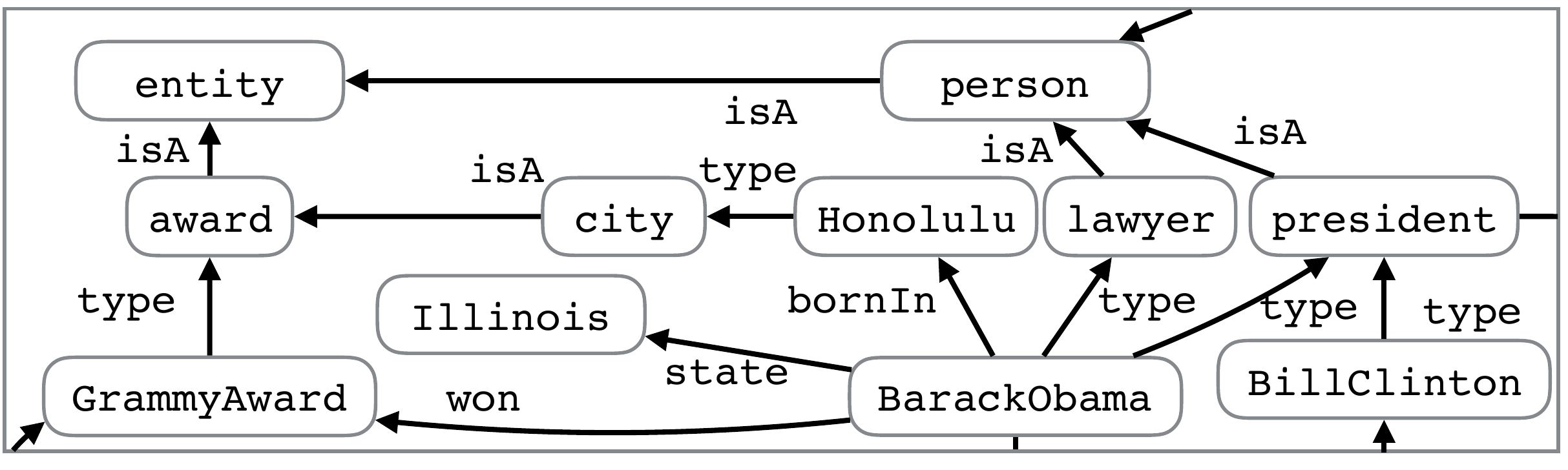} \\
 	$T_{\text{US Presidents}}=\{\struct{BarackObama}, \struct{RonaldRegan},...\}$\\
 	\vspace{1mm}
 	\hrule
 	\vspace{1mm}
 	\includegraphics[width=0.8\linewidth]{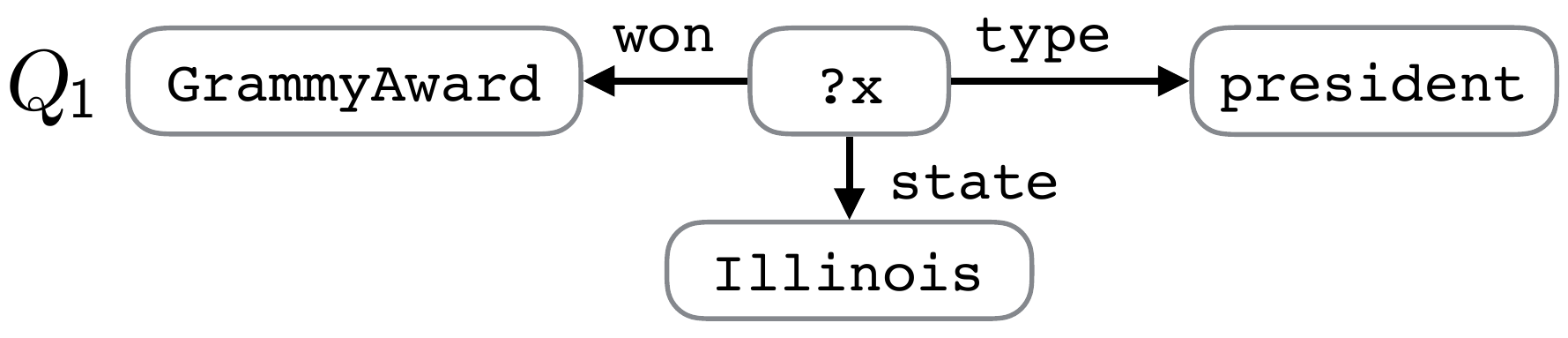}\\
 	\utterance{This president from Illinois won a Grammy.}\\
 	\raggedright
 	$\mathit{diff}(Q_1,\struct{BarackObama})=hard$ \\
 	\vspace{2mm}
 	$\mathit{dist}^{hard}_{Q_1} = \struct{RonaldRegan}$ $~~~~~~~~~$ $\mathit{dist}^{easy}_{Q_1} = \struct{HarryTruman}$
 	\caption{A fragment of a KG, a topic, and a hard question generated
 		from it. Two distractors for turning it into a multiple choice question are shown, one easy to rule out and one hard (Regan had a Hollywood career before becoming president).}
 	\label{fig:intro-example}
 	\vspace{-0.5cm}
 \end{figure}

Knowledge graphs (KGs) such as YAGO~\cite{suchanek_yago} and DBpedia~\cite{AuerBKLCI07} contain facts about real-world named
entities. They provide taxonomic knowledge, for instance, that
\struct{BarackObama} is a \struct{person} as well as a
\struct{formerSenator}. They also contain factual knowledge, for
instance, that \struct{BarackObama} is married to
\struct{MichelleObama} and was born on \struct{August 4,
  1961}. Textual knowledge captures how named entities and
their relationships are referred to in natural language, for
example, \struct{BarackObama} as \phrase{Barack H. Obama}.

Easily extensible data formats such as RDF are commonly used to store
KGs, which makes it easy to complement them with additional facts
without having to worry about a predefined schema. RDF stores facts as
(subject, predicate, object) triples, which can then be queried using
SPARQL as a simple-yet-powerful structured query language.

In this work, we address the problem of generating quiz-style
knowledge questions from KGs. 
As shown in Figure \ref{fig:intro-example}, starting from 
a KG and a topic such as \textit{US Presidents}, we generate
a quiz question whose unique answer is an entity from that topic.
The question starts its life as an automatically generated
triple-pattern query, which our system verbalizes.
Each generated question is adorned with a
difficulty level, providing an estimate for how hard it is to answer,
and optionally a set of distractors, which can be listed alongside the
correct answer to obtain a multiple-choice question. Our system is
able to judge the impact of the
distractors on the difficulty of the resulting multiple-choice question.

\textbf{Applications} of automatically generated knowledge questions
include education and evaluation. One way to educate users about a
specific domain (e.g., Sports or Politics) is to prompt them with
questions, so that they pick up facts as they try to answer --
reminiscent of flash cards used by pupils. When qualification for a
task needs to be ensured, such as knowledge about a specific domain,
automatically generated knowledge questions can serve as a
qualification test. Crowdsourcing is one concrete use case as outlined
in~\cite{DBLP:conf/websci/SeylerYBA16}. Likewise, knowledge questions
can serve as a form of CAPTCHA to exclude likely bots.

\textbf{Challenges}. To discriminate how much people know about a
domain, it is typical to ask progressively more difficult
questions. In our setting, this means that we need to automatically
quantify the difficulty of a question. This is not a trivial task as
it requires to take into consideration multiple signals and their
interaction. One might, for example, consider all questions whose
answer is \struct{BarackObama} to be easy, as he is a prominent
entity. However, very few people would know that he won a
\struct{GrammyAward}. It is therefore important to identify signals
that predict question difficulty and to combine them in a meaningful
manner.

Answers provided by the user should be easy to verify
automatically. In our setting, we want to ensure that disputes about
the correctness of an answer are minimal, since we envision a setting
with minimal human involvement (on the asking side). One important way
to achieve this is by ensuring that each question has exactly one
correct answer. Complementary to having questions with unique correct
answers is dealing with possible variation in user input (e.g.,
\phrase{Barack Obama} vs \phrase{Barack H. Obama}). One way
of overcoming this is by turning fill-in-the-blank questions to
multiple-choice questions. Here, one needs to carefully consider the
impact distractors have on question difficulty.

A final challenge is the production of well-formed natural language
questions. We are interested not only in correct language, but also in
generating questions that do not look artificial. Such questions are
desirable not only for their aesthetic appeal, but also minimize
the chance of humans discovering that such questions were generated
automatically. An important consideration here is how to ensure that
coherent questions have sufficient variety. For example, while a $KG$
may classify \struct{BarackObama} as both an \struct{entity} and a
\struct{formerSenator}, we would like to use the latter in asking
about him, as the first is unnatural. Similarly, while the relation
connecting \struct{TobeyMaguire} to \struct{Spider-Man} might be
called \struct{actedIn}, we would like to have some variety in how
this is expressed (e.g., \phrase{acted in} or \phrase{starred in})

\begin{figure}[t]
  \centering
  \includegraphics[width=1.0\linewidth]{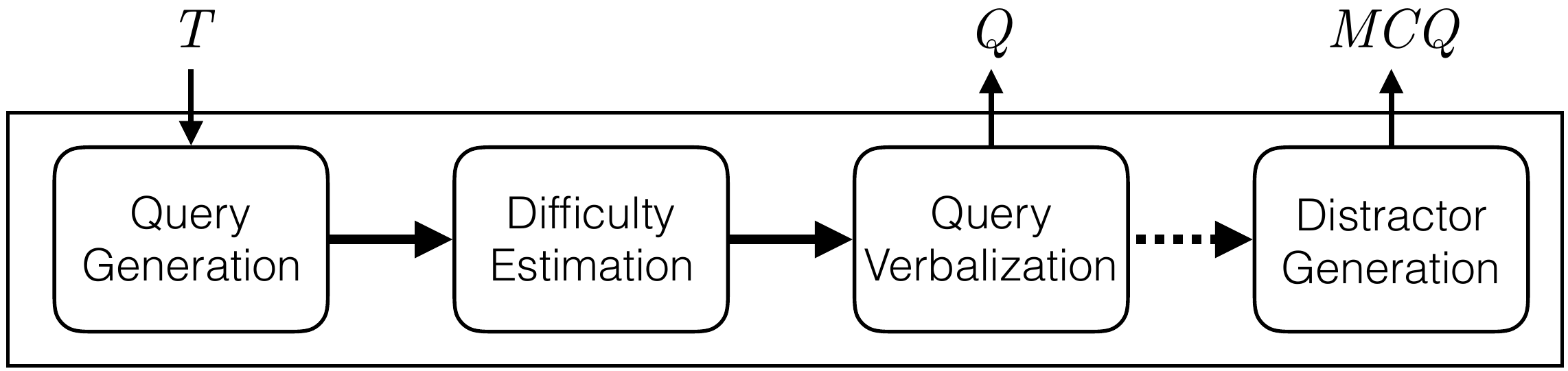}
  \caption{Question generation pipeline.}
  \label{fig:pipeline}
  \vspace{-0.5cm}
\end{figure}

\textbf{Contributions}. We propose an end-to-end approach to the novel
problem of generating quiz-style knowledge questions from knowledge
graphs. Our approach has three major components: query generation,
difficulty estimation, and query verbalization to generate a
question. In a setting where multiple-choice questions are desired, a
fourth component takes care of both generating the distractors and
quantifying their impact on question
difficulty. Figure~\ref{fig:pipeline} depicts our pipeline for
generating questions and multiple choice questions.

The query generation component generates a structured query that will
serve as the basis of the final question shown to a human. By starting
from a structured query, we are able to generate questions that are
certain to have exactly one unique, correct answer in our knowledge
graph. In query generation, several challenges need to be addressed so
that the resulting cues are meaningful.

Difficulty estimation is one of the challenges that needs to be
addressed. To estimate the difficulty of a structured query, we
leverage different signals about contained named entities, which we
derive from a Web-scale document collection annotated with named
entities from the KG. To learn weighting those signals, we make use of
more than thirty years' worth of data from the Jeopardy! quiz show.

Since our questions start their life as structured queries over the
KG, we also verbalize them by generating a
corresponding natural language question. Following earlier work on
query verbalization and natural language generation, we adopt a
template-based approach. However, we extend this approach with
automatically mined paraphrases for relations and classes in the
KG, ensuring diversity in the resulting natural language
questions.

\textbf{Outline}. The rest of this paper unfolds as
follows. Section~\ref{sec:preliminaries} introduces preliminaries and
provides a formal statement of the problem addressed in this
work. Following that, we provide details on each stage shown in
Figure~\ref{fig:pipeline}. Section~\ref{sec:query-generation}
describes how a SPARQL query can be generated that has a unique answer
in the KG. Our approach for estimating the difficulty of the generated
query is subject to
Section~\ref{sec:difficulty-estimation}. Section~\ref{sec:query-verbalization}
describes how the query can be verbalized into natural
language. Extensions for multiple-choice questions are described in
Section~\ref{sec:distractor-gen}. Section~\ref{sec:experiments} lays
out the setup and results of our experiments. We put our work in
context with existing prior research in
Section~\ref{sec:related-work}, before concluding in
Section~\ref{sec:conclusion}.
 
\section{Preliminaries and \\ Problem Statement}
\label{sec:preliminaries}

We now lay out preliminaries and formally state the problem addressed
in this work.

\textbf{Knowledge Graphs} (KGs) such as as Freebase
\cite{BollackerEPST08}, Yago \cite{suchanek_yago}, and DBpedia
\cite{AuerBKLCI07} describe \emph{entities} $E$ (e.g., \struct{BarackObama}) by connecting them to
other entities, \emph{types} $T$ --- also called classes (e.g., \struct{president}, 
\struct{leader}), and
\emph{literals} $L$ (e.g., \phrase{1985-02-05}) using
\emph{predicates} $P$ (e.g., \struct{bornIn}, \struct{birthdate}, \struct{type}). A
KG is thus a set of facts (or triples),
$\{f \mid f \in E \cup T \times P \times E \cup T \cup L\}$. A triple
can also be seen as an instance of a binary predicate, with the first
argument called the \emph{subject} and the second called the
\emph{object}, hence the name subject-predicate-object (SPO).
Figure \ref{fig:intro-example} shows a KG fragment.

Pattern-matching is used to query a KG. Given a set of variables $V$ that are always prefixed with a
question mark (e.g., \struct{?x}), a \emph{triple-pattern-query} is a
set of triple patterns $Q=$
$\{q \mid q \in V\cup E \cup T \times V\cup P \times V \cup E \cup T
\cup L\}$.
An answer $a$ to a query is a total mapping of variables to items in
the KG such that the application of $a$ to each $q$ results in a fact
in the KG. In our setting, inspired by Jeopardy!, we restrict ourselves to queries having a
single variable for which a unique answer exists in the KG. Put
differently, there exists only one binding of the single variable to a
named entity, so that all triple patterns have corresponding facts in
the KG.

More specifically, we use Yago2s \cite{suchanek2013yago2s} as our reference knowledge graph in
this work. Yago2s is automatically constructed by combining information
extraction over Wikipedia info\-boxes and categories with the lexical
database WordNet \cite{fellbaum98wordnet}. In total, Yago2s contains 2.6m entities, 300k
types organized into a type hierarchy, and more than a hundred predicates which are used to form
more than 48m facts. Yago entities are associated with Wikipedia
entries, whereas a Yago type corresponds to a WordNet synset or Wikipedia category.
To compute signals necessary for estimating
question difficulty, we make use of the ClueWeb09/12 document
collections and the FACC annotations provided by Google \cite{FACC}. The latter
provide semantic annotations of disambiguated named entities from
Freebase, which we can easily map to Yago2s via their corresponding
Wikipedia article. An annotated sentence in this corpus looks as follows:
\begin{center}
\textit{``}[\textit{Obama}|\struct{BarackObama}] \textit{endorsed} [\textit{Clinton}|\struct{HillaryClinton}] \textit{earlier today.''}
\end{center}

\textbf{Jeopardy!} is a popular U.S. TV quiz show that features
comprehensive natural language questions that are referred to as
clues. Clues are usually posed as a statement and the required answer
is in turn posed as a question. For instance, in Jeopardy! the
question: \textit{This fictional private investigator was created by
  Arthur Conan Doyle.} has the answer: \textit{Who is Sherlock
  Holmes?} Clues come with monetary values, corresponding to the
amount added to a contestant's balance when answering correctly. We
reckon that monetary values correlate with human performance and thus
question difficulty -- a hypothesis which we investigate in
Section~\ref{sec:difficulty-estimation}.

\textbf{Problem Statement}. Put formally, our objective in this work
is to automatically generate a question $\mathcal{Q}$ whose unique answer is an
entity $e \in \mathfrak{T}$ which can be supported by facts in the KG.
$\mathfrak{T}$ is a thematic set of entities called a topic \emph{topic},
which allows us to control
the domain from which knowledge questions are generated (e.g., \textit{American Politics}).
 Moreover, we assume a predefined set of \emph{difficulty
  levels} $D = \{d_1,...,d_n\}$ with a strict total order $<$ defined
over its elements, and we want to estimate the difficulty of providing the 
answer $a$ to $\mathcal{Q}$,
denoted $\mathit{diff}(\mathcal{Q}, a)$. An extension of the above problem which we
also deal with in this work is the generation of \emph{multiple choice
 questions} (MCQ's), where the task is to extend a question $\mathcal{Q}$ into
a $MCQ$ by generating a set of incorrect answers, called
\emph{distractors}, and quantifying their difficulty.

In our concrete instantiation of the above problem, we use Wikipedia
categories as topics and Yago2s as our KG. As a first attempt
to address the above problem, we consider a setting with two
difficulty levels, $D = \{easy, hard\},$ where $easy < hard$.
For our purposes, a \emph{question} is any natural language sentence
that requires an answer. It can look like what we think of as a question, 
or as a declarative sentence in the same style as Jeopardy! clues.

\section{Query Generation}
\label{sec:query-generation}

The first stage in our pipeline is the generation of a query that has a unique 
answer in the KG.
This query serves as the basis for generating a question that will be shown to
human contestants. The unique answer will be the one a contestant needs to 
provide in order to correctly answer the question. 
As is common practice in quiz-games, ensuring that a question has a single answer
 simplifies answer verification.

The input to the query generation step is a topic $\mathfrak{T}$. The unique answer to the generated 
query will be an entity 
$e \in \mathfrak{T}$ randomly drawn from the KG. Query generation is guided by the following desiderata: 
i)~the query should contain at least one type triple pattern, which is crucial when 
verbalizing the query to generate a question 
(e.g., \utterance{Which \underline{president} \ldots}), and
ii)~entities mentioned in the query should not give any obvious clues about the
answer entity. In what follows we present the challenges in achieving
each of these desiderata, and our solutions to these challenges.

\newpage
\subsection{Answer Type Selection}
Questions asking for entities always require a type that is either specified 
implicitly (e.g., \phrase{who} for \struct{person} and \phrase{where} for 
\struct{location}) or explicitly 
(e.g., \utterance{Which \underline{president} \ldots}). Here we address the problem 
of selecting a type to refer to the answer entity in the question.
KGs tend to contain a large number of types and typically associate
an entity with multiple types. Some of these types are easy for an average human
to understand and typically appear in text talking about an entity (e.g., \struct{president}, \struct{lawyer}). Other types, however, are artifacts of attempts to have an 
ontologically complete and formally sound type system. Such types are meaningful 
only in the context of a type system, but not on their own (e.g., the type 
\struct{entity} or \struct{thing}).

We use our entity-annotated corpus to capture the salience of a semantic type $t$ for an 
entity $e$, denoted $s(t,e)$. 
We start by collecting occurrences of an entity $e$ along with \emph{textual types} to 
which it belongs $t_{text}$ in our 
entity-annotated corpus. We use the following patterns
to collect $(t_{text},e)$ pairs: \\
\textbf{Pattern \#1:}\\ 
\struct{ENTITY} (\phrase{is a}$\mid$\phrase{is an}$\mid$\phrase{, a}$\mid$\phrase{and other}$\mid$\phrase{or other}) TYPE\\
``\struct{BarackObama} \textit{and other \underline{presidents} attended the ceremony.}''\\
\vspace{-2mm}\\
\textbf{Pattern \#2:}\\ 
TYPE  (\phrase{like}$\mid$\phrase{such as}$\mid$\phrase{including}$\mid$\phrase{especially}$\mid$) \struct{ENTITY}\\
``\textit{...several \underline{attorneys} including} \struct{BarackObama}''\\
These 
patterns are inspired by Hearst \cite{DBLP:conf/coling/Hearst92}.

The next step before computing semantic type salience is to disambiguate
$(t_{text},e)$ pairs to $(t,e)$ pairs --- note that entities are already disambiguated
in the corpus, so we only  need to disambiguate $t_{text}$ to a semantic type $t$
in the KG. Relying on the fact that our semantic types are WordNet synsets \cite{fellbaum98wordnet}, 
we use the lexicon that comes with WordNet (e.g., \{\textit{lawyer, attorney}\} $\rightarrow$ \struct{lawyer}) for generating a set of semantic type candidates
for a given textual type. We then use a simple yet effective heuristic where a textual type $t_{text}$ paired with an 
entity $e$ is disambiguated to a semantic type $t$ if i) $t$ is in the set of
candidates for $t_{text}$ and ii) $e \in t$.

We compute salience $s(t,e)$ as the relative frequency with which the disambiguated
$(t,e)$ pair was observed in our corpus. To select a type for the answer entity $e$, we 
draw one of the types to which it belongs randomly based on $s(t,e)$.

\subsection{Triple Pattern Generation}
We now have an answer entity $e$ and one of its semantic types $t$ that will be
used to refer to $e$ in the question. We now need to create a query (which 
includes the type constraint $t$) whose unique answer over the KG 
is $e$. We focus here on questions with unknown entities as these are the 
ones we can use Jeopardy! data to train our difficulty classifier on \cite{Ferrucci_Watson}. 
In principle, we can allow for unknown relations or types as well if we had 
the right training data.
Creating a query means selecting facts where $e$ is either the subject
or object and turning these into triple patterns by replacing $e$ with a variable
(\struct{?x}). Not all facts can be used here, as some reveal too much about the
answer and render the question too trivial. Other facts will be redundant 
given the facts already used.

\textbf{Elimination of Textual Overlap with the Answer}. 
The first restriction we impose on a fact is that the surface forms of entities
that appear in it cannot have any textual overlap with surface forms of the answer 
entity. The question \utterance{This president is married to Michelle Obama.} reveals too much about the answer entity. For
overlap, we look at the set of words in the surface forms, excluding common stop
words. We discuss our approach to collecting surface forms for entities in 
Section \ref{ssec:verbalization-lexicons} below.

\textbf{Elimination of Redundant Facts}. Given a set of facts that has been 
chosen, a new fact does not always add new information. Keeping this new fact in a 
query will result in an awkwardly phrased question that can be clearly 
identified by a human as having been automatically generated. 
In our example from Figure \ref{fig:intro-example} we decided to use the type 
\struct{president} to ask about \struct{BarackObama}. Using the fact \struct{(BarackObama type politician)} or the fact 
\struct{(BarackObama type person)} to extend the question is clearly redundant and adds no extra information. 
To eliminate this issue, we check each new type fact against all existing ones. 
If the new type is a supertype (e.g., \struct{person}) of an existing one (e.g., \struct{president}), we discard it.

\section{Difficulty Estimation}
\label{sec:difficulty-estimation}

We now describe our approach to estimating the difficulty of 
answering the knowledge query generated in Section \ref{sec:query-generation}. There are several, 
seemingly contradictory, signals that affect the difficulty of a question. 
As discussed earlier, one might expect any question asking for a popular entity such as \struct{BarackObama} 
to be an easy one. However, if we were to ask \utterance{This president from Illinois won a Grammy Award.},
few people are likely to think of \struct{BarackObama}.
We use a classification model trained on a corpus of questions paired with their 
difficulties to predict question difficulty.

Note that the difficulty is computed based on the query and not its 
verbalization, which we generate in the next section. 
Our goal here is to create questions 
that measure factual knowledge rather than linguistic ability. We elaborate on
this point further in Section \ref{sec:query-verbalization}.

Since we rely on supervised training for difficulty estimation, we make the 
natural assumption that the difficulty labels in the training and `testing' 
questions are drawn from the same underlying distribution for some target 
audience. 
We also assume that for this population, it is possibly to capture the difficulty
of a question. As evidence for this, in the Jeopardy! dataset \cite{jarchive} we find a positive 
correlation between the attempted questions for a certain difficulty-level and the 
number of times a question of this difficulty-level could not be answered. For the
five difficulty-levels (\$200, \$400, \$600, \$800, \$1000), 4.46\%, 8.35\%, 12.69\%,
17.82\% and 25.69\% of the questions could not be answered, respectively.

\subsection{Data Preparation}
\label{ssec:data-preparation}
We use the Jeopardy! quiz-game show data described in Section \ref{sec:preliminaries} for training
and testing our difficulty estimation classifier. 
The larger goal is to estimate the difficulty of answering queries generated from a 
knowledge graph, so we restrict ourselves to a subset of the Jeopardy! questions
answerable from Yago \cite{suchanek_yago}, which we collected as described below. 
However, all methods and tools are general enough to apply to a setting other 
than ours of Jeopardy!/Yago.

We say a question is answerable form Yago if i) all entities mentioned in the
question and its answer are in Yago, and ii) all relations connecting these 
entities are captured by Yago. To find these questions, we automatically 
annotate the questions with Yago entities using the Stanford CoreNLP named entity
recognizer (NER) \cite{DBLP:conf/acl/FinkelGM05} in conjunction with the AIDA 
tool for named entity disambiguation \cite{HoffartYBFPSTTW11}. We concatenate
the output of the NER system with the answer entity, which we annotate as 
an entity mention as well, and pass it to AIDA for an improved
 disambiguation context. 
An example of an input to AIDA
looks as follows:
\vspace{-1mm}
\begin{center}
\textit{[Shah Jahan] built this complex in [Agra, India] to immortalize [Mumtaz],
	his favorite wife. [Taj Mahal]} 
\end{center}
\vspace{-1mm}
and the corresponding  disambiguated output is:
\vspace{-1mm}
\begin{center}
\struct{ShahJahan} \textit{built this complex in} \struct{Agra} \textit{to immortalize} \struct{MumtazMahal}\textit{, his favorite wife.} \struct{TajMahal} 
\end{center}

We retain an entity-annotated question if i)~its answer can be mapped to a 
Yago entity, ii)~its body has at least one entity (the one that will be given in 
the question, not the answer), and iii)~considering all
entities in the question and the answer entity, each entity can be
paired with another entity to which it has a direct relation in
Yago. The last condition ensures that we have questions that can be
captured by the relationships in Yago. However, it does not identify
this relation, and such a match may be spurious. Since this is hard to
establish automatically, we invoke humans at this point.

We run a crowdsourcing task on the questions that survive the above
automated annotation and filtering procedure. The task is to assign
one of two labels to an entity-annotated question/answer pair. A
question/answer pair is to be labeled \textit{Good} if i)~all
entities in the question have been captured and disambiguated
correctly, ii)~the question can be captured by relations in
Yago, and iii)~the answer is a unique one. The crowdsourcing task ran
until we obtained a total of 500 questions that we use in our experiments.

\subsection{Difficulty Classifier}
After obtaining the data needed for training and testing a difficulty 
classifier, we turn our attention to building this classifier and the 
features used to do so. Formally, our goal is to learn a function
$\mathit{diff}(Q,e) \in \{easy, hard\}$ that learns the difficulty of providing the answer $e$
to the query $Q$.

We use logistic regression as our model of choice. We chose this
specific model due to the ease with which it can be trained and
because it allows easy inspection of feature weights, which proved
helpful during development. As we are dealing with a binary classification 
case ($easy, hard$ classification), we train our model to learn 
the probability of the question being an $easy$ one, $P(\mathit{diff}(Q,e)=easy)$ , 
and set a decision boundary at 0.5. We judge a question to be $easy$ if $P(\mathit{diff}(Q,e)=easy)>0.5$ and $hard$ otherwise.

The model, however, only works if provided with the right
features. Table~\ref{table:feature-description} provides a summary of
our features and a brief description of each. The key ingredients in our feature 
repertoire are \emph{entity salience},   per \emph{coarse semantic type} salience, and
\emph{coherence of entity pairs}.

\textbf{Entity Salience ($\phi$)} is a normalized score that is used as a proxy
for an entity's popularity. As our entities come from Wikipedia, we use the 
Wikipedia link structure to compute entity salience as the relative frequency
with which the Wikipedia entry for an entity is linked to from all other 
entries.
We also consider salience on a per-coarse-semantic-type basis. The
second group of Table~\ref{table:feature-description} defines a set of
templates. We consider the coarse semantic types \struct{person},
\struct{location}, and \struct{organization} and define a fourth
coarse semantic type \struct{other} that collects entities not in any
of the three aforementioned coarse types (e.g., movies, inventions). 
Having specialized features
for individual coarse-grained types allows us to take into
consideration some particularities of these coarse types. For example,
locations tend to have disproportionately high salience. By having a
feature that accounts for this specific semantic type, we can mitigate
this. Without this feature, having a location in a question would
result in our classifier always labeling the question as easy. 

\textbf{Coherence of entity pairs ($\varphi$)} captures the relative
tendency of two entities to appear in the same context. This feature
essentially informs us about how much the presence of one entity
indicates the presence of the other entity. For example, we would
expect that:
\begin{center}
$\varphi(\struct{\small BarackObama},\struct{\small WhiteHouse})$ >
$\varphi(\struct{\small BarackObama},\struct{\small GrammyAward})$.
\end{center}
The reason is
that the first pair is more likely to co-occur together than the
second one.
All else being equal, we would expect a question asking for
\struct{BarackObama} using the \struct{WhiteHouse} in the question to
be easier than one asking for him using \struct{GrammyAward}. Intuitively, 
coherence counteracts the effect of salience. Since \struct{BarackObama} is a salient 
entity, we would expect questions asking for him to be relatively easy. 
However, asking for him using \struct{GrammyAward} is likely to make
the question difficult, as people are unlikely to make a connection between the 
two entities.

We capture coherence using Wikipedia's link structure. Given two
entities $e_1$ and $e_2$, we define their coherence as the Jaccard coefficient of
the sets of Wikipedia entries that link to their respective entries in Wikipedia.
The intuition here is that any overlap corresponds to a mention of the relation 
between these two entities.
For the above measures, we take their maximum, minimum, average, and
sum over the question as features as detailed in Table \ref{table:feature-description}.


\begin{table}
	\begin{center}
	\begin{scriptsize}
		\def\arraystretch{1.2}
		\begin{tabular}{|l|l|}
			\hline
			\textbf{Feature} & \textbf{Description} \\
			\hline
			\multicolumn{2}{|l|}{Entity Salience}  \\ \hline
			$\phi_{target}$ & answer entity salience \\
			$\phi_{min}$ & min. salience of question entities \\
			$\phi_{max}$ & max. salience of question entities \\
			$\phi_{\Sigma}$ & sum over salience of  entities \\
			$\phi_{\mu}$ & mean salience of question and answer entities \\
			$\phi_{\mu}^{q}$ & mean salience of entities in question \\
			\hline
			\multicolumn{2}{|l|}{Per-coarse-semantic-type Salience}  \\ \hline
			$\phi_{min}^{\mathcal{T}}$ & min. salience of entities of type $\mathcal{T}$ \\
			$\phi_{max}^{\mathcal{T}}$ & max salience of entities of type  $\mathcal{T}$ \\
			$\phi_{\Sigma}^{\mathcal{T}}$ & sum over salience of entities of type  $\mathcal{T}$ \\
			$\phi_{\mu}^{\mathcal{T}}$ & mean salience of entities of type  $\mathcal{T}$ \\
			\hline
			\multicolumn{2}{|l|}{Coherence}  \\ \hline
			$\varphi_{min}$ & maximum pairwise coherence of all entity pairs \\
			$\varphi_{\Sigma}$ & sum over coherence of all entity pairs \\
			$\varphi_{\mu}$ & average coherence of all entity pairs \\ 
			$\varphi_{\mu}^{QTA}$ & average coherence of entity pairs that involve answer \\
			\hline
			\multicolumn{2}{|l|}{Answer Type}  \\ \hline
			$I_{\mathcal{T}}$ & binary indicator: answer entity is of type $\mathcal{T}$ \\
			\hline
			
		\end{tabular}
		\caption{Difficulty estimator features and their description. $\mathcal{T}$ is one of 
		\struct{person}, \struct{organization}, \struct{location},  or \struct{other}.}
		\label{table:feature-description}
			\end{scriptsize}
	\end{center}
\end{table}

\section{Query Verbalization}
\label{sec:query-verbalization}

We now turn to the problem of query verbalization, whereby we transform a query
constructed in Section \ref{sec:query-generation} into a natural language 
question. 
A human can digest this question without the technical expertise required to 
understand a triple pattern query. Our final goal is to construct well-formed 
questions that are easy to understand.

The goal of our questions is to test factual knowledge 
as opposed to linguistic ability. The way that a question is formulated is not
a
factor in predicting its difficulty. This guides our approach to query
verbalization, which ensures uniformity in how questions are phrased. 

We rely on a hand crafted \emph{verbalization template} and
automatically generated \emph{lexicons} for transforming a query into a question. 
The verbalization template specifies where the different components of the query 
appear in the question. 
The lexicon serves as a bridge between knowledge graph entries and natural 
language. We start by describing our template and then move to our lexicon 
generation process.

\subsection{Verbalization Template}
Our approach to verbalizing queries is based on templates. Such approaches are
standard in the natural language generation literature \cite{DBLP:reference/nlp/2010,Reiter:2000:BNL:331955}. We adopt
a template inspired by the Jeopardy! quiz game show given in Figure 
\ref{fig:verbalization-template}. Most of the work is done in the function 
$\mathtt{verbalize}$.

\begin{figure}[h!]
    Input: Query, $Q = \{q_1,...,q_n\}$\\

    $Q_{type}$ := $\{q_i \in Q \mid \text{has the predicate}~ \struct{type}\} = \{q_{t_1},\ldots,q_{t_m}\}$\\
    $Q_{instance}$ := $Q \setminus Q_{type}= \{q_{i_1},\ldots,q_{i_l}\}$\\

  $\mathit{This}~\mathtt{verbalize}(q_{t_1}),\ldots,~\mathit{and}~\mathtt{verbalize}(q_{t_m})\\
  ~~\mathtt{verbalize}(q_{i_1}), \ldots,~and~\mathtt{verbalize}(q_{i_l})\;.$
  
  \caption{Verbalization Template}
  \label{fig:verbalization-template}
\end{figure}

The function $\mathtt{verbalize}$ takes a triple pattern and produces its verbalization.
How this verbalization is performed depends on the nature of the triple pattern.
More concretely, there are three distinct patterns possible in our setting (see
Section \ref{sec:query-generation}):
\squishlist
  \item \textbf{Type}: if the predicate is \struct{type}, then this results 
        in verbalizing the object, which is a semantic type.
  \item \textbf{PO}: where the triple pattern is of the form \struct{?var p o} and 
  \struct{p} is not \struct{type}.
  \item \textbf{SP}:  where the triple pattern is of the form \struct{s p ?var} and 
  \struct{p} is not \struct{type}.
\squishend
By considering these cases individually we ensure that linguistically well-formed
verbalizations are created. Figure \ref{fig:tp-verbalization} shows an example
of each of the three cases above. Verbalizing a triple pattern requires that
we are able to verbalize its constituent semantic items (entities, types, and
predicates) in a manner that is considerate of the specific pattern. We
present our solution to this next.

\begin{figure}
   \centering
   \begin{small}
  \begin{tabular}{| l | l | l |}
    \hline
    \textbf{Triple Pattern}       & \textbf{Pattern}   & \textbf{Verbalization} \\ \hline
    \struct{?x type movie}        & type               & \phrase{film}, \phrase{movie} \\ \hline
    \struct{?x actedIn Heat}      & PO                 & \phrase{acted in the movie Heat}\\
                                  &                    & \phrase{starred in the film Heat}  \\ \hline
    \struct{AlPacino actedIn ?x}  & SP                 & \phrase{Al Pacino appeared in}\\ 
     \hline
  \end{tabular}   
  \end{small}
 \caption{Examples of results of $verbalize(q)$.}
 \label{fig:tp-verbalization}
\end{figure}

\subsection{Verbalization Lexicons}
\label{ssec:verbalization-lexicons}
Semantic items in the knowledge graph are simply identifiers that are not 
meant for direct human consumption. It is therefore important that we map 
each semantic item to phrases that can be used to represent it 
in a natural language string such as a question.

\textbf{Entities.} To verbalize entities we follow the approach of Hoffart et 
al. \cite{HoffartYBFPSTTW11} and rely on the fact that our entities come from Wikipedia. 
We resort to Wikipedia for extracting surface forms of our entities. For each 
entity $e$, we collect the surface forms of all links to $e$'s Wikipedia entry.
We consider this text to be a possible
verbalization of $e$. 

The above process extracts many spurious verbalizations of an entity $e$. 
To overcome this issue, we associate with each candidate verbalization the 
number of times it was used to link to $e$'s Wikipedia entry and restrict 
ourselves to the five most frequent ones, which we add to the lexicon for the entry
corresponding to $e$.

\textbf{Predicates.} As Figure \ref{fig:tp-verbalization} shows, predicate 
verbalization depends on the the pattern in which it is observed (SP or PO).
We rely on our large entity-annotated corpus described in Section \ref{sec:preliminaries} for mining
predicate verbalizations sensitive to the SP and PO patterns. For each triple 
$(e_1~p~e_2) \in KG$, we collect all sentences in our corpus that match the 
patterns $Pat_{SP}=$\utterance{$e_1$ $w_1$...$w_n$ $e_2$} (e.g., ``\struct{BarackObama} \textit{was born in} \struct{Hawaii}'') and
 $Pat_{PO}=$\utterance{$e_2$ $w_1$...$w_n$ $e_1$} (e.g., ``\struct{Hawaii} \textit{is the birthplace of} \struct{BarackOmaba}'') . Following the distant supervision assumption  
 \cite{DBLP:conf/acl/MintzBSJ09}, we
hypothesize that \phrase{$w_1...w_n$} is expressing $p$. 
The above hypothesis does not always hold. To filter out possible noise we 
resort to a combination of heuristic filtering and scoring. We remove from the 
above verbalization candidate set any phrases that are longer than 50 characters
or contain a third entity $e_3$. We subsequently score how good of a fit a phrase 
\phrase{$w_1...w_n$} is for a predicate $p$ using normalized pointwise mutual 
information (npmi). For each predicate $p$, we retain the 5 highest scoring
verbalizations for each of the two patterns, $Pat_{SP}$ and $Pat_{PO}$, which 
are used for verbalizing SP and PO triple patterns, respectively.

\textbf{Types.} As explained in Section \ref{sec:preliminaries}, our types are WordNet synsets. 
We therefore rely on the lexicon distributed as part of WordNet for type paraphrasing.

Each of the three lexicons provides several ways to verbalize a semantic
item. When verbalizing a specific semantic item, we choose a verbalization
uniformly at random to ensure variety.

\section{multiple-choice questions}
\label{sec:distractor-gen}

The final component in our question generation framework turns a
question into a multiple-choice question. This has several advantages: in
general, it is easier to administer a multiple-choice question as the
problem of answer verification can be completely mechanized. This is
particularly true in cases where questions are not administered though
a computer, where such things as completion suggestion can ensure
canonical answers. In general, where knowledge questions are involved
(as opposed to free response questions that might involve opinion),
the use of multiple-choice questions is widespread as observed in such
tests as the GRE.

Turning a question into multiple-choice requires \emph{distractors}:
entities that are presented to the user as answer candidates, but are
in fact incorrect answers. Of course, not all entities constitute
reasonable distractors. A negative example would be entities that
are completely unrelated to the question. In addition to being 
related to the question, distractors should ideally be related to the 
correct answer entity. It should
generally be possible to confuse a distractor with the correct answer
to make a multiple-choice question interesting. We call this the
\emph{confusability} of a distractor. The more confusable a distractor
is with the correct answer, the more likely a test taker is to choose
it as an answer, making the multiple-choice question more challenging.

In what follows we take a look at the problems of generating
distractors in our framework and quantifying the confusability of
these distractors.

\subsection{Distractor Generation}
\label{sec:distractor-generation}

Our starting point for generating distractors is the query
$Q=\{q_1,...,q_n\}$ generated in Section~\ref{sec:query-generation},
which formed the basis of the question verbalized in
Section~\ref{sec:query-verbalization}. By starting with a query, we
have a fairly simple but powerful scheme for generating
distractors. By removing one or more triple patterns from $Q$ we
obtain a query $Q' \subset Q$ that has more than one answer entity.
All but one of these entities are an incorrect answer to $Q$.

The relaxation scheme described above can generate a large number of
candidate distractors. However, not all relaxations stay close to the
original query. If a relaxation deviates too much from $Q$, the obtained
distractors become meaningless. We address this by imposing two
restrictions on relaxed queries used to generate distractors: (i)~a
semantic type restriction, and (ii)~a relaxation distance restriction.

Semantic type restriction ensures that the answer and distractor are
type-compatible. For example, a multiple-choice question asking for a
location should not have a person as one of its distractors. The
semantic type restriction requires that a semantic type triple pattern
$Q$ is relaxed to the corresponding coarse type.

The relaxation distance restriction refers to relaxations involving
instance triple patterns. We define the distance between a query $Q$
and a query $Q' \subseteq Q$ as follows:
\[dist(Q,Q') = |\mathit{answers}(Q')| - |\mathit{answers}(Q)|,\]
where $\mathit{answers}(Q')$ is the set of answers of $Q'$
($|\mathit{answers}(Q)|$ is always 1).  We restrict relaxed queries to
have a distance of no more than $\alpha$, which we set to 10. By
pooling the results of all relaxed queries, we form a set of candidate
distractors. The choice of distractor is based on how much difficulty
we want the distractors to introduce using our notion of distractor
confusability.

\subsection{Distractor Confusability}
\label{ssec:distractor-confusability}

All things equal, a multiple-choice question can be made more or less
difficult by the choice of distractors. If one of the distractors is
highly confusable with the answer entity, the multiple-choice question
is difficult. If none of the distractors is easy to confuse with the
answer entity, the multiple-choice question is easy.

Based on this observation we regard a distractor as confusable if it
is likely to be the answer to the original question based on our 
difficulty model. This implies that if an entity is very likely to 
be the answer to a question asking about a different entity, this entity pair must
be similar. We can therefore define confusability between 
the question's answer $e_a$ and a distractor entity $e_{dist}$ as follows:
\begin{gather*}
\mathit{conf}(Q, e_a,e_{dist}) =\\ 1 - | P(\mathit{diff}(Q,e_a)=easy) - P(\mathit{diff}(Q,e_{dist})=easy) |.
\end{gather*}

Since we can have more than one distractor in a multiple-choice question, we
capture the above intuition regarding how multiple distractors affect
the overall difficulty of the question. We observe that a multiple-choice 
question is as confusing as its most confusing distractor and define the confusability
of a distractor set $\mathit{Dist} = \{e_{dist1}, e_{dist2},...\}$ as:
\[\mathit{conf}(Q,e_a, Dist) = \operatorname*{max}_{e_{dist} \in \mathit{Dist}} \mathit{conf}(Q,e_a,e_{dist}).\]

Looking at the big picture, we relate the notion of confusability in a multiple-choice question with our 
earlier notion of difficulty by combining $\mathit{diff}(Q,e_a) \in \{easy, hard\}$ and 
$\mathit{conf}(Q,e_a, Dist) \in [0,1]$ as shown in Table \ref{table:diff-conf}.
We see that an easy question can be turned in two a hard one when a very confusable distractor
is added, since the user has to distinguish between two very similar entities. However, adding an easy distractor to a hard question will not change its difficulty because even when both entities are not similar to each other, the user still has to know which entitiy is the correct answer.

\begin{table}
	\centering
	\begin{small}
		\begin{tabular}{|l||l|l|}
			\hline
			& \boldmath$diff(Q,e_a)$ & \boldmath$diff(Q,e_a)$ \\
			& \boldmath$= easy$       & \boldmath$= hard$ \\ \hline \hline
			\boldmath$conf(Q,e_a,Dist) < 0.5$	&  	easy              &   hard  \\ \hline
			\boldmath$conf(Q,e_a,Dist) > 0.5$	&   hard 			  &   hard    \\ \hline
		\end{tabular}
		
	\end{small}
	
	\caption{Combining question difficulty and multiple-choice question confusability
		into an overall difficulty in a multipe-choice setting.}
	\label{table:diff-conf}
	\vspace{-0.5cm}
\end{table}

\section{Experimental Evaluation}\label{sec:experiments}
In the following section we evaluate our approach to knowledge question generation from knowledge graphs. We perform two user studies which focus on evaluating the difficulty model and our distractor generation framework.

\subsection{Human Assessment of Difficulty}
\label{ssec:human-assessment-of-difficulty}
An important motivation for automating difficulty assessment of questions is the
fact that it is difficulty to judge for the average human what constitutes an easy 
or hard question. Beinborn et al. \cite{DBLP:journals/tacl/BeinbornZG14} has 
already shown this result for
language proficiency tests, where language teachers were shown to be bad at 
predicting the difficulty of questions when considering the actual performance 
of students. We would like to observe if the same applies to our setting. 
To create fair and informative tests, it is crucial that we are able to 
correctly assess the difficulty of a question. 

We start with the assumption that the creators of Jeopardy! are good at 
automatically assessing question difficulty. Evidence for this was discussed in
Section \ref{sec:difficulty-estimation}, where we showed that there exists a
correlation between the monetary value of a question and the likelihood of it
being incorrectly answered by Jeopardy! contestants.

In our experiment we want to show how well the average human can predict the difficulty of  a question. To do so, we randomly sampled 100 easy (\$200) and 100 hard (\$1000) questions from the 500 questions generated in Section \ref{sec:difficulty-estimation} to maximize the discrepancy in question difficulty. We then asked three human evaluators ($eval_{1}$, $eval_{2}$, $eval_{3}$) to annotate each of the 200 questions as easy or hard. We then compared their answers with 
each other and with the ground truth according to Jeopardy!.

Table \ref{table:evaluators-agreement} shows the agreement between each pair of 
human evaluators and the majority vote difficulty assessment using 
Fleiss' Kappa \cite{fleiss1971}. When looking at pairwise agreement between 
evaluators, it  ranges from fair to moderate \cite{landis1977}. This leads us
to conclude that it is hard for non-experts to properly judge the difficulty
of questions.

We also compared the majority vote of the evaluators on the difficulty of the 
questions with the ground truth provided by Jeopardy!. The result was agreement
on 62.5\% of questions. This suggests that there is a need to automate the task.

\begin{table}
	\begin{center}
		\begin{tabular}{|c||c|c|c|}
			\hline
			& $eval_{2}$ & $eval_{3}$ & majority \\
			\hline \hline
			$eval_{1}$ & 0.192 & 0.325 & 0.500 \\
			\hline
			$eval_{2}$ &  & 0.443 & 0.661 \\
			\hline
			$eval_{3}$ & & & 0.810 \\
			\hline
		\end{tabular}
		\caption{Agreement between human evaluators (all measurements are Fleiss' Kappa)}
		\label{table:evaluators-agreement}
		\vspace{-0.3cm}
	\end{center}
\end{table}

\subsection{Question Difficulty Classification}
We start by looking at the quality of our scheme for assigning difficulty levels
to questions. The scheme is described in Section \ref{sec:difficulty-estimation}, where the possible difficulty levels are $D=\{easy, hard\}$. We train our logistic regression 
classifier on 500 Jeopardy! questions annotated as described in Section \ref{sec:difficulty-estimation}.
Using ten-fold cross validation, our classifier was able to correctly identify the
difficulty levels of questions with an accuracy of 66.4\%.

To gain insight into how informative our features are, we performed a feature
ablation study where we look at the results for all combinations of our features. For this part, we grouped our features into three classes:
\squishlist
 \item \textbf{SAL:} ``Salience'' features as in Table \ref{table:feature-description}, with additional log-transformation of salience values to deal with long-tail entities.
 \item \textbf{COH:} ``Coherence'' features in Table \ref{table:feature-description}.
 \item \textbf{TYPE:} ``Per-coarse-semantic-type Salience'' and ``Answer Type'' features in Table \ref{table:feature-description}.
\squishend

Table \ref{table:features-performance} shows the results of this experiment. 
Each row corresponds to a certain combination of features enabled or disabled. Rows are shown in descending order of ten-fold cross validation accuracy. It can be seen that best performance is achieved when all of our features are integrated. From this observation it can be reasoned that all features are necessary and give complementary signals. The bottom row corresponds to a random classifier. 

\begin{table}
	\begin{center}
		\begin{tabular}{|c|c|c||c|}
			\hline
			 \textbf{SAL} & \textbf{COH} & \textbf{TYPE} & \textbf{Accuracy} \\
			\hline \hline
			 yes & yes & yes & \textbf{66.4\%} \\
			\hline
			 yes & no & yes & 65.8\% \\
			\hline
			 yes & yes & no & 62.6\% \\
			\hline
			 yes & no & no & 62.2\% \\
			\hline
			 no & no & yes & 60.0\% \\
			\hline
			 no & yes & yes & 57.8\% \\
			\hline
			 no & yes & no & 52.4\% \\
			\hline
			 no & no & no & 50.0\% \\
			\hline
		\end{tabular}
		\caption{Ablation study results for features introduced in Section \ref{sec:difficulty-estimation}. Accuracy is based on ten-fold cross-validation of the difficulty classifier's predictions.}
		\label{table:features-performance}
		\vspace{-0.7cm}
	\end{center}
\end{table}

\subsection{User Study on Difficulty Estimation}

In the following we perform an experiment on how well our classifier agrees with relative difficulty assessments of humans for questions generated by our system. It is important to note that we ask humans
for relative difficulty assessments as opposed to absolute difficulties, since we have shown in Section \ref{ssec:human-assessment-of-difficulty} that humans are not very proficient in judging absolute difficulties.

For the user study we sampled a set of $50$ entities with at least $5$ 
non-\struct{type} facts in Yago. For each entity, we generated a set
of three questions and presented them with the answer entity
to human annotators. The annotators were asked order these questions by their relative  difficulty and were allowed to skip a set of questions about 
an entity if they were not familiar with the entity. 

We then compared the correlation between the ranking given by each of the human 
annotators and the output of our logistic regression classifier. For this
we used Kendall's $\tau$, which ranges from -1, in the case of perfect disagreement,
to 1, in the case of perfect agreement. 

A total of 13 evaluators took part in the study and evaluated 92.5 questions on average. Rankings produced by the difficulty classifier moderately agree with the human annotators with $\tau=0.563$. When the $\tau$-values for users are weighted by study
participation, the average rises to  $\overline{\tau}=0.593$. Here, each user's contribution to the final average depends on how many questions she evaluated to avoid overly representing users that evaluated only few questions.

\subsection{Distractors Confusability}

We now turn to the evaluation of distractor generation for multiple-choice 
questions. Our goal is to accurately predict the confusability
of a distractor given a question's correct answer. In Section \ref{ssec:distractor-confusability} we 
presented our scheme for quantifying distractor confusability and how it fits
into a multiple-choice question setting. We evaluate our
approach here.  

For this experiment we automatically generate 10,000 ~mul\-ti\-ple-choice questions.
Each question has three answer ~choi\-ces, which are the correct answer and two distractors. 
We then restricted ourselves to 400 multiple-choice questions whose distractor pair has 
the largest difference in confusability. This was done to maximize the probability that 
study participants can actually discriminate the more confusable from the less
confusable distractor. 

We ran each multiple-choice question through a crowdsourcing platform and asked 
workers to judge which distractor is more confusing. Each multiple-choice
question was judged by 5 workers so we could take the majority vote in case the
judgments where not unanimous. 
We then compare this majority vote with the result of our confusability 
estimator. Our estimator agreed with the human annotations on 76\% of the 
400 multiple-choice questions. This translates to a Cohen's $\kappa$ of 
0.521, indicating moderate agreement \cite{cohen1960}.

\section{Related Work} 
\label{sec:related-work}

There has been work on knowledge question generation for testing linguistic 
knowledge and reading comprehension. The generation of language proficiency 
tests has been tackled in several works \cite{DBLP:conf/aaaifs/Gates11,NarendraAs13,
sakaguchi_discriminative_2013}. Here, the focus is on generating cloze
(fill-in-the-blank) tests. Beinborn et al. \cite{DBLP:journals/tacl/BeinbornZG14}
presents an approach for predicting the difficulty of answering such questions 
with multiple blanks using SVMs trained on four classes of features that 
look at individual blanks, their candidate answers, their dependence on other
blanks, and the overall question difficulty. 

Question generation for reading comprehension is aimed at evaluating 
knowledge from text corpora. This includes including general Wikipedia knowledge
\cite{DBLP:conf/premi/BhatiaKS13,Heilman09questiongeneration} and specialized
domain such as medical texts \cite{Agarwal:2011:AGQ:2043132.2043139,DBLP:conf/icwl/WangHL07}. 
While the above works focus on generating a question from a single document,
Questimator \cite{DBLP:conf/ijcai/GuoKKBB16} generates multiple choice  
questions from the textual Wikipedia corpus by considering multiple documents
related to a single topic to produce a question.
Work in this area 
has mostly taken the approach of overgeneration and ranking 
\cite{Heilman09questiongeneration,DBLP:conf/icwl/WangHL07}.  Multiple questions are generated for a 
given passage using rules. A learned model ranks the questions in terms of 
``acceptability''. In this setting, acceptable answers should be sensical, 
grammatical, and their answers should not be obvious. 

Recent work has started to look at the problem of generating questions, 
including multiple choice ones, from KGs and ontologies \cite{DBLP:conf/owled/AlsubaitPS14,DBLP:conf/www/SeylerYB15,song2016domain,DBLP:conf/acl/SerbanGGACCB16}.
Strong motivations for studying this problem, compared to question generation from text,
 are scenarios where structured data is what is available at hand, and the ability
 to generate deeper, structurally more complex questions. 
 Our system is an end-to-end solution for this problem over a large
 KG.

In Section \ref{sec:query-verbalization} we presented a simple approach for 
query verbalization that sits 
our needs. The query verbalization problem has been tackled by
Ngomo et al. for SPARQL \cite{ngonga_ngomo_sorry_2013,Ell12spartiqulation}, and Koutrika et al.
for SQL \cite{DBLP:conf/icde/KoutrikaSI10}, with a 
focus on usability. Similar to our approach, these earlier works take a 
template-based approach to verbalization, which are very widely used on the
natural language generation from logical form such as SPARQL queries 
\cite{DBLP:reference/nlp/2010,Reiter:2000:BNL:331955}.

Much recent work has focused on keyword search \cite{DBLP:conf/semweb/BlancoMV11} and question answering, rather than generation, 
from knowledge graphs \cite{DBLP:conf/cikm/BastH15,DBLP:conf/ijcai/CuiXW16,DBLP:conf/ekaw/LopezNSUMd10,DBLP:conf/www/ShekarpourNA13,DBLP:conf/www/UngerBLNGC12,DBLP:conf/aaai/XuZFHZ15,DBLP:conf/sigmod/ZouHWYHZ14}, possibly in 
combination with textual data \cite{DBLP:journals/ftir/BastBH16,DBLP:conf/sigir/SavenkovA16,DBLP:conf/cikm/YinDKBZ15}. The value of knowledge graphs is that they return crisp
answers and allow for complex constraint to answer structurally complex questions.
Of course, question answering has a long history, with one of the major highlights
being IBM's Watson \cite{DBLP:journals/ibmrd/Ferrucci12}, which won the Jeopardy! game show 
combining both structured and unstructured sources for answering.

One important contribution of our work is an approach to compute the 
difficulty of questions generated. This topic has received attention lately
in community question answering \cite{DBLP:conf/emnlp/LiuWLH13,DBLP:conf/emnlp/WangLWG14}, 
by using a competition-based approach that tries to capture how much skill
a question requires for answering. There has also been work on estimating
query difficulty in the context of information retrieval \cite{DBLP:series/synthesis/2010Carmel,DBLP:conf/sigir/Yom-TovFCD05} to learn an estimator that predicts the expected precision of the 
query by analyzing the overlap between the results of the full query and the 
results of its sub-queries.

\section{Conclusion}
\label{sec:conclusion}

We proposed an end-to-end approach to the novel problem of generating
quiz-style knowledge questions from knowledge graphs. Our approach
addresses the challenges inherent to this problem, most importantly
estimating the difficulty of generated questions. To this end, we
engineer suitable features and train a model of question difficulty on
historical data from the Jeopardy! quiz show, which is shown to
outperform humans on this difficult task. A working prototype
implementing our approach is accessible at:
\begin{center}
  \url{https://gate.d5.mpi-inf.mpg.de/q2g}
\end{center}

\newpage

\bibliographystyle{abbrv}
\bibliography{k-questions}

\begin{thebibliography}{10}

\bibitem{jarchive}
J! {Archive}.
\newblock \url{http://j-archive.com}.

\bibitem{Agarwal:2011:AGQ:2043132.2043139}
M.~Agarwal and P.~Mannem.
\newblock Automatic gap-fill question generation from text books.
\newblock In {\em BEA}, 2011.

\bibitem{DBLP:conf/owled/AlsubaitPS14}
T.~Alsubait et~al.
\newblock Generating multiple choice questions from ontologies: Lessons learnt.
\newblock In {\em {OWLED}}, 2014.

\bibitem{AuerBKLCI07}
S.~Auer et~al.
\newblock {DBpedia: A Nucleus for a Web of Open Data}.
\newblock In {\em {ISWC}/{ASWC}}, 2007.

\bibitem{DBLP:journals/ftir/BastBH16}
H.~Bast et~al.
\newblock {Semantic Search on Text and Knowledge Bases}.
\newblock {\em Foundations and Trends in IR}, 10(2-3), 2016.

\bibitem{DBLP:conf/cikm/BastH15}
H.~Bast and E.~Haussmann.
\newblock {More Accurate Question Answering on Freebase}.
\newblock In {\em {CIKM}}, 2015.

\bibitem{DBLP:journals/tacl/BeinbornZG14}
L.~Beinborn et~al.
\newblock {Predicting the Difficulty of Language Proficiency Tests}.
\newblock {\em {TACL}}, 2, 2014.

\bibitem{DBLP:conf/premi/BhatiaKS13}
A.~S. Bhatia et~al.
\newblock Automatic generation of multiple choice questions using wikipedia.
\newblock In {\em PReMI}, 2013.

\bibitem{DBLP:conf/semweb/BlancoMV11}
R.~Blanco et~al.
\newblock Effective and efficient entity search in {RDF} data.
\newblock In {\em ISWC}, 2011.

\bibitem{BollackerEPST08}
K.~D. Bollacker et~al.
\newblock {Freebase: a Collaboratively Created Graph Database for Structuring
  Human Knowledge}.
\newblock In {\em {SIGMOD}}, 2008.

\bibitem{DBLP:series/synthesis/2010Carmel}
D.~Carmel and E.~Yom{-}Tov.
\newblock {\em Estimating the Query Difficulty for Information Retrieval}.
\newblock Morgan {\&} Claypool Publishers, 2010.

\bibitem{cohen1960}
J.~Cohen.
\newblock {A Coefficient of Agreement for Nominal Scales}.
\newblock {\em Educational and Psychological Measurement}, 20(1):37, 1960.

\bibitem{DBLP:conf/ijcai/CuiXW16}
W.~Cui et~al.
\newblock {KBQA: an Online Template Based Question Answering System over
  Freebase}.
\newblock In {\em {IJCAI}}, 2016.

\bibitem{Ell12spartiqulation}
B.~Ell et~al.
\newblock {Spartiqulation -- Verbalizing SPARQL Queries}.
\newblock In {\em ILD Workshop, ESWC}, 2012.

\bibitem{fellbaum98wordnet}
C.~Fellbaum, editor.
\newblock {\em {WordNet: an Electronic Lexical Database}}.
\newblock MIT Press, 1998.

\bibitem{DBLP:journals/ibmrd/Ferrucci12}
D.~A. Ferrucci.
\newblock Introduction to "this is watson".
\newblock {\em {IBM} Journal of Research and Development}, 2012.

\bibitem{Ferrucci_Watson}
D.~A. Ferrucci et~al.
\newblock {Building Watson: An Overview of the DeepQA Project}.
\newblock {\em {AI} Magazine}, 31(3), 2010.

\bibitem{DBLP:conf/acl/FinkelGM05}
J.~R. Finkel et~al.
\newblock {Incorporating Non-local Information into Information Extraction
  Systems by Gibbs Sampling}.
\newblock In {\em {ACL}}, 2005.

\bibitem{fleiss1971}
J.~L. Fleiss.
\newblock {Measuring Nominal Scale Agreement among Many Raters.}
\newblock {\em Psychological Bulletin}, 1971.

\bibitem{FACC}
E.~Gabrilovich et~al.
\newblock {FACC1}: Freebase annotation of {ClueWeb} corpora, {Version} 1, 2013.

\bibitem{DBLP:conf/aaaifs/Gates11}
D.~M. Gates.
\newblock {How to Generate Cloze Questions from Definitions: {A} Syntactic
  Approach}.
\newblock In {\em {AAAI}}, 2011.

\bibitem{DBLP:conf/ijcai/GuoKKBB16}
Q.~Guo et~al.
\newblock {Questimator: Generating Knowledge Assessments for Arbitrary Topics}.
\newblock In {\em {IJCAI}}, 2016.

\bibitem{DBLP:conf/coling/Hearst92}
M.~A. Hearst.
\newblock {Automatic Acquisition of Hyponyms from Large Text Corpora}.
\newblock In {\em {COLING}}, 1992.

\bibitem{Heilman09questiongeneration}
M.~Heilman and N.~A. Smith.
\newblock {Question Generation via Overgenerating Transformations and Ranking}.
\newblock Technical report, 2009.

\bibitem{HoffartYBFPSTTW11}
J.~Hoffart et~al.
\newblock {Robust Disambiguation of Named Entities in Text}.
\newblock In {\em {EMNLP}}, 2011.

\bibitem{DBLP:reference/nlp/2010}
N.~Indurkhya and F.~J. Damerau, editors.
\newblock {\em Handbook of Natural Language Processing}.
\newblock Chapman and Hall/CRC, 2010.

\bibitem{DBLP:conf/icde/KoutrikaSI10}
G.~Koutrika et~al.
\newblock {Explaining Structured Queries in Natural Language}.
\newblock In {\em {ICDE}}, 2010.

\bibitem{landis1977}
J.~R. Landis and G.~G. Koch.
\newblock {The Measurement of Observer Agreement for Categorical Data}.
\newblock {\em Biometrics, Vol. 33}, 1977.

\bibitem{DBLP:conf/emnlp/LiuWLH13}
J.~Liu et~al.
\newblock Question difficulty estimation in community question answering
  services.
\newblock In {\em EMNLP}, 2013.

\bibitem{DBLP:conf/ekaw/LopezNSUMd10}
V.~L{\'{o}}pez et~al.
\newblock Scaling up question-answering to linked data.
\newblock In {\em EKAW}, 2010.

\bibitem{DBLP:conf/acl/MintzBSJ09}
M.~Mintz et~al.
\newblock Distant supervision for relation extraction without labeled data.
\newblock In {\em {ACL}}, 2009.

\bibitem{NarendraAs13}
A.~Narendra et~al.
\newblock {Automatic Cloze-Questions Generation}.
\newblock In {\em {RANLP}}, 2013.

\bibitem{ngonga_ngomo_sorry_2013}
A.-C. Ngonga~Ngomo et~al.
\newblock Sorry, {I} {Don}'{T} {Speak} {SPARQL}: {Translating} {SPARQL}
  {Queries} into {Natural} {Language}.
\newblock In {\em {WWW}}, 2013.

\bibitem{Reiter:2000:BNL:331955}
E.~Reiter and R.~Dale.
\newblock {\em Building Natural Language Generation Systems}.
\newblock Cambridge University Press, 2000.

\bibitem{sakaguchi_discriminative_2013}
K.~Sakaguchi et~al.
\newblock {Discriminative Approach to Fill-in-the-Blank Quiz Generation for
  Language Learners}.
\newblock In {\em {ACL}}, 2013.

\bibitem{DBLP:conf/sigir/SavenkovA16}
D.~Savenkov and E.~Agichtein.
\newblock When a knowledge base is not enough: Question answering over
  knowledge bases with external text data.
\newblock In {\em {SIGIR}}, 2016.

\bibitem{DBLP:conf/acl/SerbanGGACCB16}
I.~V. Serban et~al.
\newblock Generating factoid questions with recurrent neural networks: The 30m
  factoid question-answer corpus.
\newblock In {\em {ACL}}, 2016.

\bibitem{DBLP:conf/www/SeylerYB15}
D.~Seyler et~al.
\newblock Generating quiz questions from knowledge graphs.
\newblock In {\em {WWW}}, 2015.

\bibitem{DBLP:conf/websci/SeylerYBA16}
D.~Seyler et~al.
\newblock Automated question generation for quality control in human
  computation tasks.
\newblock In {\em {WebSci}}, 2016.

\bibitem{DBLP:conf/www/ShekarpourNA13}
S.~Shekarpour et~al.
\newblock Question answering on interlinked data.
\newblock In {\em {WWW}}, 2013.

\bibitem{song2016domain}
L.~Song and L.~Zhao.
\newblock Domain-specific question generation from a knowledge base.
\newblock {\em arXiv}, 2016.

\bibitem{suchanek_yago}
F.~M. Suchanek et~al.
\newblock Yago: {A} {Core} of {Semantic} {Knowledge}.
\newblock In {\em {WWW}}, 2007.

\bibitem{suchanek2013yago2s}
F.~M. Suchanek et~al.
\newblock Yago2s: Modular high-quality information extraction with an
  application to flight planning.
\newblock In {\em BTW}, volume 214, 2013.

\bibitem{DBLP:conf/www/UngerBLNGC12}
C.~Unger et~al.
\newblock Template-based question answering over {RDF} data.
\newblock In {\em WWW}, 2012.

\bibitem{DBLP:conf/emnlp/WangLWG14}
Q.~Wang et~al.
\newblock A regularized competition model for question difficulty estimation in
  community question answering services.
\newblock In {\em EMNLP}, 2014.

\bibitem{DBLP:conf/icwl/WangHL07}
W.~Wang et~al.
\newblock Automatic question generation for learning evaluation in medicine.
\newblock In {\em {ICWL}}, 2007.

\bibitem{DBLP:conf/aaai/XuZFHZ15}
K.~Xu et~al.
\newblock {What Is the Longest River in the USA? Semantic Parsing for
  Aggregation Questions}.
\newblock In {\em {AAAI}}, 2015.

\bibitem{DBLP:conf/cikm/YinDKBZ15}
P.~Yin et~al.
\newblock {Answering Questions with Complex Semantic Constraints on Open
  Knowledge Bases}.
\newblock In {\em {CIKM}}, 2015.

\bibitem{DBLP:conf/sigir/Yom-TovFCD05}
E.~Yom{-}Tov et~al.
\newblock Learning to estimate query difficulty: including applications to
  missing content detection and distributed information retrieval.
\newblock In {\em {SIGIR}}, 2005.

\bibitem{DBLP:conf/sigmod/ZouHWYHZ14}
L.~Zou et~al.
\newblock Natural language question answering over {RDF:} a graph data driven
  approach.
\newblock In {\em {SIGMOD}}, 2014.

\end{thebibliography}

%
%
%
%
%
%
%
%

\end{document}